\documentclass{sig-alternate}
\usepackage{float}
\usepackage{array}
\usepackage{makeidx}
\usepackage{moreverb}
\usepackage[boxed,commentsnumbered]{algorithm2e}
\usepackage[colorlinks, citecolor=green]{hyperref}
\usepackage{bm}

\begin{document}
%


\CopyrightYear{2016} 
\setcopyright{acmlicensed}
\conferenceinfo{ICMR'16,}{June 06 - 09, 2016, New York, NY, USA}
\isbn{978-1-4503-4359-6/16/06}\acmPrice{\$15.00}
\doi{http://dx.doi.org/10.1145/2911996.2912014}
\clubpenalty=10000 
\widowpenalty = 10000

\title{ACD: Action Concept Discovery from\\ Image-Sentence Corpora}
%
%
%
%
%

\numberofauthors{3}
%
\author{
%
%
Jiyang Gao\\
       \affaddr{Univ. of Southern California}\\
      \email{jiyangga@usc.edu}
\alignauthor
Chen Sun\\
       \affaddr{Univ. of Southern California}\\
       \email{chensun@usc.edu}
\alignauthor Ram Nevatia\\
       \affaddr{Univ. of Southern California}\\
       \email{nevatia@usc.edu}
}



\maketitle
\begin{abstract}
Action classification in still images is an important task in computer vision. It is challenging as the appearances of actions may vary depending on their context (\emph{e.g.} associated objects). Manually labeling of context information would be time consuming and difficult to scale up. To address this challenge, we propose a method to automatically discover and cluster action concepts, and learn their classifiers from weakly supervised image-sentence corpora. It obtains candidate action concepts by extracting verb-object pairs from sentences and verifies their visualness with the associated images. Candidate action concepts are then clustered by using a multi-modal representation with image embeddings from deep convolutional networks and text embeddings from word2vec. More than one hundred human action concept classifiers are learned from the Flickr 30k dataset with no additional human effort and promising classification results are obtained. We further apply the AdaBoost algorithm to automatically select and combine relevant action concepts given an action query. Promising results have been shown on the PASCAL VOC 2012 action classification benchmark, which has zero overlap with Flickr30k.
\end{abstract}

 \category{I.5}{Pattern Recognition}{Computer Vision}

\terms{Algorithms, Experimentation, Performance}

\keywords{Concept Discovery, Multi-modal Representation, Action Classification, Domain Transfer}

\section{Introduction}

Action classification in still images is an important computer vision task with wide applications. One major difficulty for this task is that the appearances of actions usually depend not only on the action itself but also on the objects that they are applied to. For example, the action \textit{play musical instruments} is general, its appearance varies significantly when the object is different (\emph{e.g.} \textit{play violin} and \textit{play piano}); varying objects is sometimes also termed as varying context. To make the situation worse, the potential space of action and context combinations is very large.

Most previous methods in action classification~\cite{yang2010recognizing, oquab2014learning, gkioxari2014r, gkioxari2014actions, maji2011action, khosla2014integrating} focus on learning action classifiers from fully annotated datasets with limited categories~\cite{andriluka20142d,everingham2014pascal, yao2011human}. To extend the action vocabulary and allow context awareness, such methods require labeling actions in images manually, which is time-consuming and difficult to scale up. On the other hand, there are many large-scale datasets with image and sentence descriptions~\cite{ordonez2011im2text, chua2009nus, young2014image, hodosh2013framing, lin2014microsoft, krishnavisualgenome} readily available. Many of the verb-object (VO) pairs in these sentence descriptions can be treated as actions with context. However, there are two major challenges of learning action classifiers from such image-sentence corpora:

\begin{itemize}
\item \textbf{Annotation noise:} a sentence description may contain several VO pairs, many of which may not correspond to actual actions in the image.

\item \textbf{Language diversity:} humans might refer the same action category by different terms. For example, the actions of jump, leap and flip are visually similar, but there is only one tag for this kind of action, \textit{jump}, in PASCAL VOC 2012 action classification dataset.
\end{itemize}

\begin{figure}[t]
\centering
\includegraphics[scale=0.58]{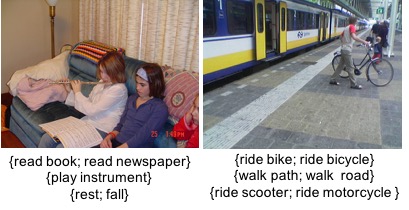}
\caption{Top 3 action recognition outputs on PASCAL VOC 2012 using our discovered action concept classifiers from Flickr30k }
\label{fig:teaser}
\end{figure}

To address these challenges, we propose a method to automatically discover \textit{action concepts}, cluster them and learn classifiers for them from image-sentence corpora (ACD). Some action recognition results are shown in Figure 1. To cope with annotation noise, the ACD framework filters action concepts based on a visually discriminative criterion: each action concept classifier must perform reliably on validation data. For the language diversity problem, ACD merges different VO pairs referring to the same actions together. 

The overall pipeline proceeds as follows: we first extract subject, verb and object triples from the sentences in image-sentence corpora with an off-the-shelf NLP parser, and filter out the ones whose subjects are non-human. We then extract a 4096-dimensional feature vector for each image by passing it through pre-trained (on ImageNet ) convolutional neural networks (CNN). Two-fold cross validation is applied to verify the visual discriminative power of action concepts. For each action concept, we construct a multi-modal representation by concatenating CNN feature vectors of its associated images and its word2vec~\cite{mikolov2013distributed} embedding features. The similarity between two action concepts is defined as the cosine similarity of the multi-modal representations. To merge different concepts referring to the same action, we cluster the discovered action concepts with a nearest neighbor (NN) clustering algorithm based on the multi-modal similarity metric. Finally, for each cluster, we train a linear SVM classifier on all the associated images. 

For our experiments, we learned action concepts from the Flickr30k \cite{young2014image} dataset which contains 30 thousand images and each image is associated with 5 sentences to describe it. The sentences are generated by Amazon Mechanical Turk. An image-sentence example is shown in Figure 2. Besides testing the classifiers on subsets of the Flickr30k dataset, to compare with other weakly supervised methods, we also test the classifiers on an independent and widely used PASCAL VOC 2012 action classification dataset, which has 10 pre-defined categories. Note that this amounts to a transfer learning task with Flickr30K as the \emph{source domain} and PASCAL as the \emph{transfer domain}. Since a certain action tag in the transfer domain could be related to multiple concept clusters, we use the action tag as search keywords to find related clusters in the cluster pool and combine all the related cluster classifiers using AdaBoost~\cite{freund1999short} for this action tag to build a stronger classifier.

The key contributions of this paper are:

1.~A methodology to discover, filter and cluster action concepts from image-sentence corpora automatically.

2.~A multimodal representation for actions by combining the convolutional neural networks feature of related images and word2vec feature of the concept terms (verb+object).


3.~A method for building a stronger classifier for transfer domain by using an ensemble of learned action classifiers from source domain.

\section{Related Work}
\textbf{Visual Concept Learning:}
There have been two trends in learning visual concept automatically. One relies on Internet search engines~\cite{chen2013neil,divvala2014learning} to collect images and train concept classifiers on those images. In particular, NEIL~\cite{chen2013neil} uses a never-ending-learning algorithm to gather visual data from internet and build relationships between these concepts. However, the vocabulary dataset is fixed. LEVAN~\cite{divvala2014learning} discovers comprehensive vocabulary from Internet data and trains detectors for them using the images collected by search engine. 

The second trend discovers concepts from weakly labelled data. Specifically, VCD~\cite{sun2015automatic} proposes a concept mining and clustering algorithm based on visual and semantic similarity. However, it mainly focuses on general object concept learning. ConceptLearner~\cite{zhou2014conceptlearner} designs a scalable max-margin algorithm to discover and to learn visual concepts from weakly labeled image dataset~\cite{ordonez2011im2text, chua2009nus}. However, it lacks concept clustering. 

\textbf{Multimodal Integration:}
There is large literature on multimodal representation models. We only select a few representative systems to describe here. \cite{kiela2014learning} adopts a concatenation strategy and use a convolutional neural network to extract features from images and skip-gram model for text. \cite{bruni2014multimodal} constructs text feature vectors and image feature vectors separately and then mixes them by Singular Value Decomposition on their concatenation. \cite{lazaridou2015combining} extends skip-gram model by taking visual information into account, which is called multimodal skip-gram. Image dispersion-based filtering in \cite{kiela2014improving} improves multi-modal representations by approximating conceptual concreteness from images and filtering model input.

\textbf{Action Classification in Still Images:}
The use of convolutional neural network (CNN) has brought huge improvement in action classification. \cite{oquab2014learning} finetunes the CNN pre-trained on ImageNet and shows improvement over traditional methods. \cite{gkioxari2014r} designs a multi-task (person-detection, pose-estimation and action classification) model based on R-CNN. \cite{gkioxari2014actions} develops a part-based approach by leveraging convolutional network features for action and attribute classification. They show top performance on PASCAL VOC human attribute and action classification.  \cite{gkioxari2015contextual} develops an end-to-end deep convolutional neural network that utilizes contextual information of actions. Most previously leading methods in action classification for 2D images are based on part detectors. In particular, \cite{maji2011action} trains action specific poselets and for each instance creates a poselet activation vector which is being classified using SVMs. In \cite{khosla2014integrating}, image regions are identified in different classes by using a dense sampling space and a random forest algorithm with discriminative classifiers. These works are all based on fully-supervised learning method and trained on fully-labeled data.

\section{Action Concept Learning}
In this section, we first present our method to extract action concepts from sentences in the source domain (Flickr30k) and verify the visualness of concepts. Secondly, our multimodal representation for action concepts and NN-clustering algorithm are demonstrated. Finally, we present the method using AdaBoost to combine multiple related action classifiers.
\begin{figure}
\centering
\includegraphics[scale=0.42]{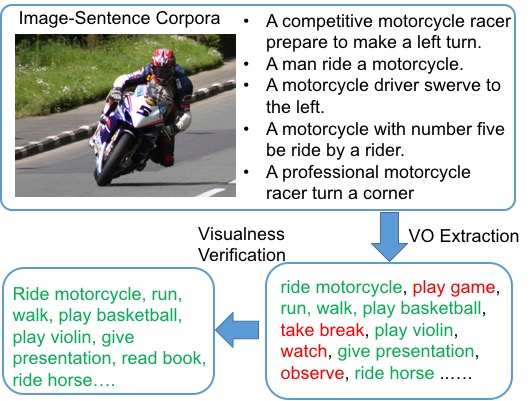}
\caption{VO pairs extraction and verification}
\end{figure}

\subsection{Action Concept Extraction}
We use grammatical relations called \emph{dependencies} to discover action concepts from sentences. Specifically, the sentences are parsed by the Stanford CoreNLP parser \cite{manning2014stanford}, and then verb-object(VO) pairs are extracted from the parsed sentences with only human subjects. A VO pair, such as play basketball, is used as action, for example ``play basketball", ``ride horse". For the actions that have no object, we use a word ``none" as their object, such as ``run none" and ``walk none". We remove the terms that occur fewer than k times (we set k=30 in our experiments). Remaining VOs are action concept candidates. Now each image may contain several VOs (\emph{i.e.} action concepts) and each VO may be associated with several images.

\subsection{Concept Visualness Verification}
After filtering VO pairs from the corpus, the remaining ones are meaningful action concept candidates, as shown in Figure 2. However, some of them may not be visually distinct. For example, there are no obvious visual patterns for the action  ``take break". Concept visualness verification is used to eliminate the visually non-distinct action concepts. We collect positive samples from the images associated with a concept and randomly choose equal number of negative samples from other images. Each image is processed by an ImageNet pre-trained VGG-16 \cite{simonyan2014very}. The network takes in a 224*224 pixel RGB image and generate a 4096-dimensional vector as the  image feature. Linear-SVM classifiers are trained for each action concept candidate and their performance is evaluated by two fold cross-validation. We prune the action concepts whose average precision (AP) is lower then 70\%.

\subsection{Concept Multimodal Representation}

For visual representation, similar to \cite{kiela2014learning}, our approach makes use of the collection of images associated with the action concept. After we extract image-level holistic features from a pre-trained deep CNN for each image, two approaches are considered to aggregate the image features for each action concept, as shown in Figure 3:

1.~Compute the average for all feature vectors (CNN-Mean).

2.~~Compute dimension-wise maximum for all feature vectors (CNN-Max).

\begin{figure} 
\centering
\includegraphics[scale=0.5]{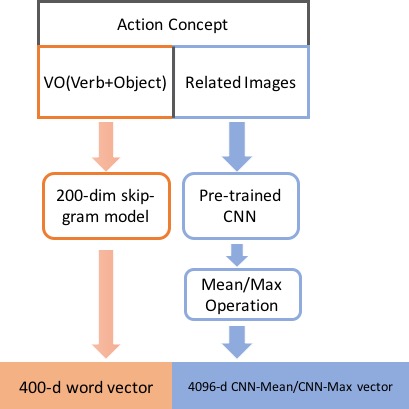}
\caption{Multimodal representation by concatenating the visual feature and linguistic feature. For the visual part, we use pre-trained CNN to extract image-level features and combine them by computing the average or component-maximum. For the linguistic part, we use word2vec to extract a 200-dimensional semantic feature vectors for ``verb" part and ``object" part and concatenate them together.}
\end{figure}

For linguistic representation, we extract 200-dimensional word embedding for each word. In particular, we train a skip-gram \cite{mikolov2013distributed} model using the English Wikipedia dump. The feature vector for the fake object ``none" is an all-zero vector. The semantic representation for an action concept is 400-dimensional (200d for the verb part and 200d for the object part).  We construct our multimodal representation for action concept by concatenating the visual representation and linguistic representation. Note that the visual feature vector and the linguistic feature vector are both L2-normalized. 
\begin{displaymath}
V_{concept}=\alpha V_{visual} \oplus (1-\alpha)V_{linguistic}
\end{displaymath}
where $\oplus $ is the concatenation operator and $\alpha$ is a parameter balancing the weights given to the visual and linguistic representations. Similarity between two concepts, $c_i$ and $c_j$, can be calculated by cosine similarity: 
\begin{displaymath}
S(c_i,c_j)=\frac{V_{c_i}V_{c_j}}{\|V_{c_i}\|\|V_{c_j}\|}
\end{displaymath}

\subsection{NN Clustering}
The total number of the action concepts is smaller then the number of general concepts, so there may be more differences and less general patterns between different datasets. To be adaptive to different situations, we make use of the patterns within the dataset to cluster the action concepts. For example, if the whole concept set is \{play basketball, play sport, play instrument\}, then there should be two clusters \{play basketball, play sport\} and \{play instrument\}. In another situation, the whole concept set is \{run, play sport, play instrument\}, then there should be three clusters. So the intuition of the solution is that the compactness of cluster should be determined by the dataset itself. 

We design a clustering algorithm based on the nearest neighbors (NN) of each concept, called NN-clustering. It's a non-parametric clustering algorithm, unlike K-means or spectral clustering which specifies the number of clusters explicitly. Suppose that the size of a cluster is k, \emph{i.e.} the number of concepts in that cluster, then this cluster meet two conditions:

1. For any two concepts, $c_i$ and $c_j$, in this cluster, suppose $c_i$ is the  nearest $n$ neighbor of  $c_j$ and $c_j$ is the  $m$ nearest neighbor of $c_i$, then $m<=k+C, n<=k+C$ , $C$ is a constant.

2. For any two concepts  $c_i$ and $c_j$  in the cluster, the similarity between $c_i$ and $c_j$,  $S(c_i,c_j)$ must be larger then the average concept similarity$S_{avg}$. $S_{avg}=\frac{1}{n^2}\sum_{i=1}^{n}\sum_{j=1}^{n} S(c_i,c_j)$, where n is the size of the whole concept set.
\begin{algorithm}
  \SetAlgoLined
  \KwData{concept similarity matrix of size $l \times l$ and concept list of size $l$}
  \KwResult{cluster list }
\tcp*[h]{initialization process: }\\
 \For{ $i \leftarrow 0 $ to $l$}{
\For{$j \leftarrow i+1$ to $l$}{
\If{concepts[i] and concepts[j] are mutually the nearest neighbour}{
	clusters.add(cluster(concepts[i],concepts[j]))\;
	}
}
} 
\tcp*[h]{iteratively clustering:}\\
  \While{unclustered concept $\neq \varnothing $}{
	change $\leftarrow$ True\;
    \While{$change==True$}{
	change $\leftarrow$ False\;
	\For {$i \leftarrow 0 $ to $l$}{
	 \For{$k \leftarrow 0$ to length(clusters)}{
	
	 \If{checkClustered(clusters[k],concepts[i])} {
	  \tcp*[h]{if concept could be added into cluster}\\
	 clusters[k].add(concepts[i])\;
	 change $\leftarrow$ True
	 }
	 
	 }
	}
	}
	index=randint(length(unclustered concepts))\;
	clusters.add(cluster(concepts[index]))\;
  
    }
  \caption{NN-Clustering Algorithm}
\end{algorithm}

\begin{figure*}
\centering
\includegraphics[scale=0.5]{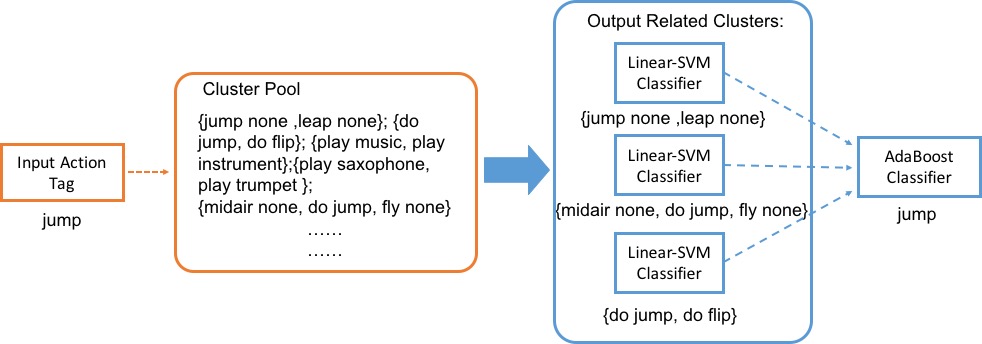}
\caption{Combining multiple classifiers of related clusters. For example, if the input action tag is ``jump", we first search for the related clusters in the concept cluster pool. The classifiers of related clusters are regarded as weak classifiers to the action tag. Then we train an AdaBoost classifier to combine all the weak classifiers to a stronger classifier for the input action tag, \emph{i.e.} jump.}
\end{figure*}

The first condition causes mutually similar concepts to be clustered together. The constant $C$ controls the compactness of the cluster. The larger is $C$, the looser is the cluster. Note that mutual similarity is a very important constraint. Suppose that we have an action concept which is not similar to any other concepts, but there is still a ranking list of its similar concepts. If we only use unilateral similarity, then it would be clustered with its most similar concept. Only the first condition would tend to make the cluster as large as possible. If we regard the whole action concept set as a cluster, then it will still satisfy the first condition. So we must use the second condition as an additional constraint.  We implement this algorithm in an iterative way, as shown in Algorithm 1.

\subsection{Training Action Classifier with SVM and AdaBoost}

After clustering concepts, we train a linear-SVM classifier for each cluster using CNN image features. We notice that action tags in traditional action classification dataset can relate to multiple concept clusters. For example, the action tag ``jump" (in PASCAL VOC 2012 action classification) relates to\emph{\{do jump, do flip\}}, \emph{\{leap none, jump none\}} and \emph{\{midair none, do jump, fly none\}}. Each related cluster can be trained as a weak classifier to the action tag and these weak classifiers can be combined to a single strong classifier using ensemble learning methods, such as AdaBoost. We search all the related clusters automatically for a specific action tag. Specifically, we use the action tag as a keyword to search in our cluster pool. If there is a concept of the cluster that matches with the action tag, then this cluster is picked out as a related action cluster. The pipeline is shown in Figure 4.

AdaBoost generates a new hypothesis by applying the weak learning algorithms to a sample of the training data.The key idea is to set weights to the training data so that wrongly classified data points are more likely to be considered by the new hypothesis. The output is given by 
\begin{displaymath}
f_T(x)=\sum_{i=1}^T\beta_i h_i(x)
\end{displaymath}
where $x$ is the input feature, $f_T(x)$ is the AdaBoost classifier, $h_i(x)$ are the linear-SVM classifier of the related clusters and $\beta_i$ is the learned goodness weight for each weak classifier.

\section{Evaluation}
In this section, we evaluate our multimodal representation with different $\alpha$ values. NN-Clustering algorithm is applied with different $C$ values. We also evaluate our action classifiers on the \emph{transfer domain} (PASCAL VOC 2012 action classification dataset) and \emph{source domain} (Flickr30k).

\subsection{Multimodal Representation}
We use all images and associated sentences in the source domain (Flickr30k) to discover action concepts. After parsing each sentence with Stanford CoreNLP and extracting VO from the whole corpus, we get 327 action concept candidates. 171 VOs pass the visualness verification step and are action concepts. 

We calculate two versions of multimodal representation (CNN-Mean and CNN-Max) for each action concept with $\alpha \in\{0, 0.1, 0.2, ..., 0.9,1.0\}$ and cluster these action concepts with $C=4$. Given the clustering results, we train linear-SVM classifiers using LIBLINEAR \cite{fan2008liblinear} for each cluster. A typical objective of clustering algorithm is to attain high intra-cluster similarity and low inter-cluster similarity, which implies that if we train classifiers for each cluster, then the performance of those classifiers should depend on the quality of the clustering results. If we fix the clustering algorithm itself, then a more powerful representation of the action concepts should yield a better clustering results, hence better classification performance. We define the average classification accuracy as the mean of accuracies of all action classifiers and use it as our metric to evaluate the multimodal representation.

\begin{figure}
\centering
\includegraphics[scale=0.3]{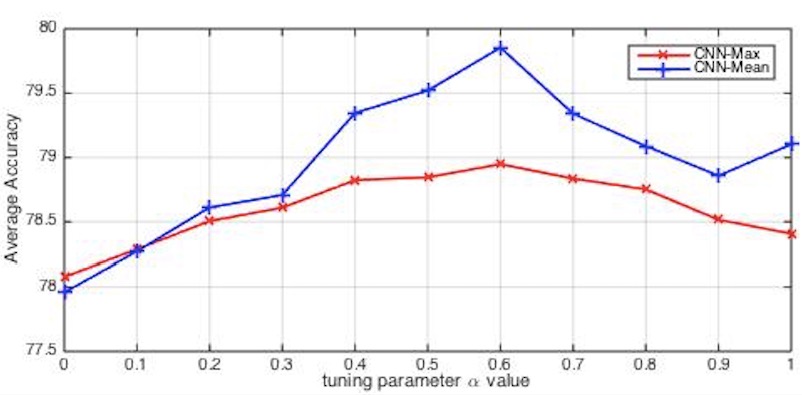}
\caption{Evaluation of the multimodal representation (CNN-Mean and CNN-Max) with different $\alpha$ values on average accuracy}
\end{figure}

From the experimental results shown in Figure 5, we can see the following: a) the performance of CNN-Mean is generally better than that of CNN-Max. The reason could be that the maximum operation is more subject to noise. b) When $\alpha= 0.6$, both method achieve the highest average classification accuracy. The parameter $\alpha$ weights the relative contribution of the linguistic part and the visual part. The larger the tuning parameter $\alpha$ is, the larger contribution the visual part will make. This observation shows that the visual part and the linguistic part are almost equally important to construct a good representation.

We discuss the influence of tuning parameter $\alpha$ in more specific examples and see the obvious differences of large $\alpha$ and small $\alpha$. When $\alpha=0.9$ (almost all visual features), some good results include \emph{\{ride bike, ride bicycle, bike none, biking none\}},  \emph{\{ride skateboard, skateboarding none, skate none\}, \{throw ball, play baseball\}}. Apparently, ``throw ball'' and ``play baseball'' are semantically-unrelated, at least not as similar as ``throw ball" and ``hit ball", but they are very similar visually (Baseball game is about throwing ball and catching ball, so when we use a large $\alpha$ value, these two action concepts are clustered together. If we adopt a small $\alpha$ value, ``throw ball" would be clustered with ``hit ball" and ``kick ball".) On the other hand, large $\alpha$ also generates some ``bad" results:  \emph{\{play trumpet, play saxophone\}}, \emph{ \{play instrument, play music\}} and \emph{\{play accordion\}}. Perhaps all these three clusters should be in one cluster, because they are all related to ``play instrument". When $\alpha=0.1$, the algorithm indeed clusters all these together:  \emph{\{play trumpet, play saxophone, play violin, play keyboard, play drum, play guitar, play music, play accordion, play instrument\}}. However, small $\alpha$ also produces poor clustering results in other cases, for example,  \emph{\{ride bike, ride bicycle, ride motorcycle, ride skateboard, ride horse, ride scooter\}}; all these actions are about riding something. According to the discussion above, both small $\alpha$ and large $\alpha$ could produce good clustering results depending on our desired goodness criterion. 

We use multiple $\alpha$ values to clustering concepts to get a comprehensive cluster set. Specifically, we calculate the CNN-Mean action representations and cluster the action concepts with $\alpha \in\{0, 0.1, 0.2, ..., 0.9,1.0\}$. The union of all these clusters from different $\alpha$ values make a comprehensive cluster pool.  


\subsection{Clustering Algorithm}
\begin{figure}
\centering
\includegraphics[scale=0.35]{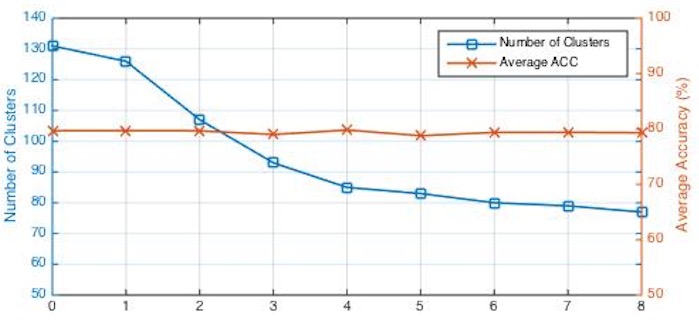}
\caption{NN-clustering evaluation by changing the value of the constant $C$.}
\end{figure}

We evaluate the NN-clustering algorithm with different value of constant $C$, where $C$ controls the compactness of the cluster. The larger the value of $C$, the looser the cluster will be, that is the concepts in the same cluster will be less similar. We test the algorithm with $C \in \{0, 1, 2, 3, 4, 5, 6, 7, 8\}$ on average classification accuracy and the number of clusters, as shown in Figure 6. The average classification accuracy is the same as shown in Section 4.1.

According to the experimental results, we make two observations: a) the average accuracy doesn't change much with constant $C$; b) the number of the clusters is stable around 80 when $C>4$. Based on these observations, we fix $C=4$ and $\alpha=0.6$. Some cluster examples are shown in Table 1.

\begin{table}[H]
\centering

\begin{tabular}{|l|m{5.5cm}|l|} \hline
Index&Cluster Terms&AP(\%)\\ \hline
1&\{play trumpet, play saxophone, play violin, play guitar, play accordion, play instrument, hold guitar, play keyboard, play drum\} &68.13\\  [0.5 ex]\hline
2&\{play soccer, play hockey, play football, play sport, play basketball, play baseball, play tennis, play volleyball, compete none\}&71.38\\  [0.5 ex]\hline
3& \{climb mountain, climb face, climb wall, climb rock, climbing none\}&80.05 \\ [0.5 ex] \hline
4& \{bike none, biking none\}&69.55 \\  [0.5 ex]\hline
5& \{ride bike, ride bicycle, ride motorcycle, ride scooter\}& 86.92\\  [0.5 ex]\hline
6&\{read book, read newspaper\}& 75.71 \\  [0.5 ex]\hline
7& \{cooking none, cook none, kitchen none\}&84.51 \\  [0.5 ex]\hline
8& \{prepare food, prepare meal, eat food, drink beer, have drink\}& 52.24\\ [0.5 ex] \hline
9& \{give presentation, give speech\}&67.28\\  [0.5 ex]\hline
10& \{walk sidewalk, walk street, cross street\}&52.62\\ [0.5 ex] \hline
11& \{swim trunk, swim none, pool none, diving none\}& 65.78\\  [0.5 ex]\hline
12&\{kayake none, paddle none\}&71.85 \\  [0.5 ex]\hline
13& \{surf none, surf wave\}& 90.33\\ [0.5 ex] \hline
14& \{hold baby, hold boy, hold infant\}& 62.46\\ [0.5 ex] \hline
15& \{kick ball, throw ball, hit ball\}& 58.91\\ [0.5 ex] \hline
16& \{perform trick, do trick\} &58.56\\ [0.5 ex] \hline
17& \{sing none, singing none\}& 73.32\\ [0.5 ex] \hline
18& \{ski none\} &87.60\\ [0.5 ex] \hline
19& \{leap none, jump none\} &58.65\\ [0.5 ex] \hline
20& \{take photo, take photograph, take picture, hold camera\} &66.82\\ [0.5 ex] \hline
&Average of all 84 clusters&65.48 \\ [0.5 ex] \hline
\end{tabular}
\caption{Concept cluster examples are selected based on the number of positive samples in each and sorted in descending order. The AP of random guess is 10\%}
\end{table}

\begin{table*}\small
\centering
\begin{tabular}{|l|l|l|l|*{10}{l}|r|}

\hline
method & supv.& domain&box &jump&phone&instr.& read &bike &horse&run& photo& comp.& walk &mAP\\ \hline
NO-ACD& weak&transfer&no&38.1&12.2 &76.2 & 12.9 &83.5&83.5&24.9&22.4 & 68.9 &19.6&44.2\\
ACD($\alpha=0.6$)& weak&transfer&no&\textbf{63.4}&15.4 &71.7 & 28.9 &66.9&69.7&53.5&13.9 & 69.2 &30.9&48.3\\
ACD-A &weak&transfer&no&62.2&\textbf{15.4} &\textbf{78.8} &\textbf{29.6} & \textbf{84.5} &\textbf{85.9}& \textbf{60.8}& \textbf{24.0}& \textbf{69.2}& \textbf{32.4}& \textbf{54.3}\\ \hline
VGG-SVM &full&same&no&73.1 &43.1&84.8 & 45.4& 86.7 &92.0 &73.1 &34.5 &76.0&30.7 & 63.9\\
Oquab \textit{et al} \cite{oquab2014learning} &full&same&yes&74.8 &46.0&75.6 & 45.3& 93.5 &95.0 &86.5 &49.3 &66.7&69.5 & 70.2\\
\hline\end{tabular}
\caption{Comparison of different methods on the PASCAL VOC 2012 action classification test set (AP\%). ``Box" denotes the bounding box in the training data of PASCAL VOC 2012.  Except Oquab \textit{et al.}\cite{oquab2014learning}, other methods don't use any bounding box information to train. The abbreviations in table: ``instr."= ``play instrument", ``bike"=``ride bike", ``horse"=``ride horse", ``photo"=``take photo", ``comp."=``use computer". The AP of ``use computer" and ``phone" are the same among our three methods, that's because only one cluster for this action is discovered by our algorithm.}
\end{table*}

\begin{figure*}
\centering
\includegraphics[scale=0.6]{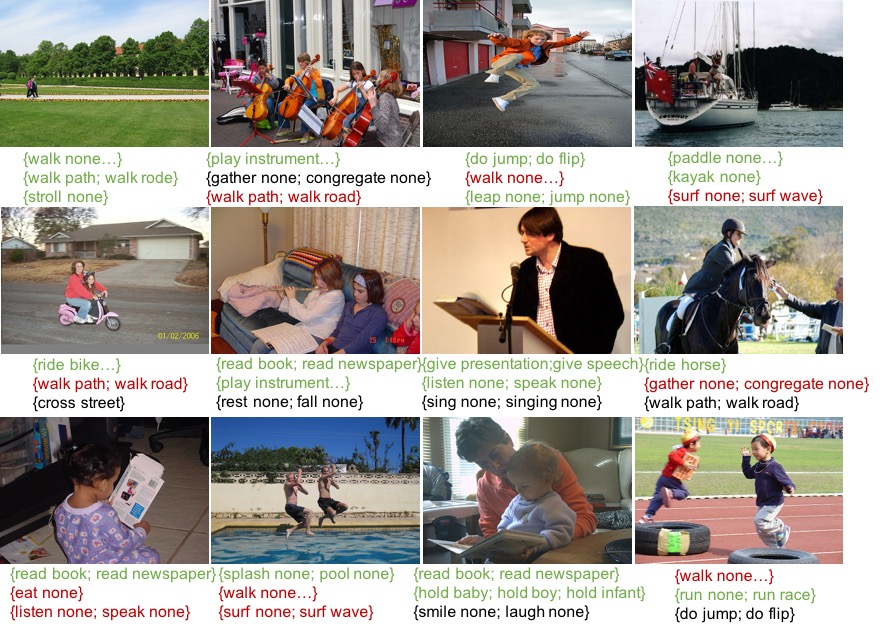}
\caption{Some examples of the top 3 prediction results on PASCAL VOC 2012 using the discovered action classifiers in $\text{ACD}(\alpha=0.6)$. Green represents correct prediction, red represent wrong prediction and black stands between correct and wrong prediction. \{ride bike...\} stands for \{ride bike, ride bicycle, bike none, biking none\}; \{walk none...\} stands for \{walk none, walk street, walk dog, walk sidewalk, walk past\}; \{paddle none...\} stands for \{paddle none; paddle boat; boat none; fishing none\}; \{play instrument...\} stands for\{play keyboard, play guitar, play instrument, play violin, play drum, play music, play accordion\}.  }
\end{figure*}

\subsection{Action Classification}
\textbf{Data:} ACD discovers and learns action concepts from \emph{source domain}, which is Flickr30k. Flickr30k contains 30 thousand images and each image is associated with 5 sentences to describe it. The sentences are generated by Amazon Mechanical Turk. PASCAL VOC 2012 action classification is used as \emph{transfer domain} to test the discovered action concepts. It's a standard benchmark for action classification in still images and there are 10 action tags in it: \{jump, phone, play instrument, read, ride bike, ride horse, run, take photo, use computer, walk\}.

\textbf{Metric:} Average precision (AP) and mean average precision (mAP) are used as evaluation metric in the action classification task.

First we evaluate the classification performance on the source domain. The associated images of each cluster are split into two halves: one half is used for training and the other half is used for testing. The negative samples are randomly selected from other clusters and the ratio of negative samples and positive samples is 10:1. There are 84 clusters in total; 20 of them are shown in Table 1. Concept cluster examples in Table 1 are selected based on the number of positive samples in each and sorted in descending order. The mAP of all 84 cluster classifiers is 65.48\%. The random mAP is 10\%. Note that the mAP can be boosted by reducing the number of concept clusters but at the cost of losing specificity. The AP is lower on some clusters, such as \{prepare food...\} and \{walk...\}, because they have larger intra-class variance and don't have particular patterns. For example, ``walk" is a similar action to ``run". On the other hand, ``walk" contains many different varieties, such as ``walk on the street" and ``walk on the grass", and those varieties have different scenes.

We further evaluate our action classifiers on PASCAL VOC 2012 test server and results are shown in Table 2.  Note that we don't use any data from PASCAL to finetune the action classifiers learned from the source domain (Flickr30k).  The baseline method (denoted as \textbf{NO-ACD}) is trained as follows: use pre-defined vocabularies (\emph{i.e.} action tags in PASCAL VOC 2012 action classification dataset) as search keywords, collect image data in Flickr30k that associated sentences contain those tags and train linear-SVM classifiers for each action tag using the associated images.
For the method \textbf{ACD-A}, we search in the classifier pool to find all the related action clusters and train an AdaBoost classifier for each action tag. The training data for the AdaBoost classifier is the union of images of each related cluster from Flickr30k (no PASCAL data is used). We also train a linear SVM classifier on the  \emph{trainval} set of PASCAL VOC 2012 action classification as a reference (denoted as \textbf{VGG-SVM} in Table 2). Oquab \textit{et al} \cite{oquab2014learning} finetunes the pre-trained CNN \cite{krizhevsky2012imagenet} on PASCAL VOC 2012 action classification dataset. 

The item ``box" list in Table 2 denotes whether the method makes use of the persons' bounding boxes in PASCAL VOC 2012. The item ``supv." in Table 2 provides the kind of supervision adopted by the method. The item ``domain" in Table 2 provides whether there is domain transfer between training set and testing set. Some action recognition results on PASCAL VOC are shown in Figure 7. 

\textbf{Weak Supervision and Domain Transfer:} Our action concepts are discovered and learned from image-sentence corpora, which is a weakly-labelled dataset. We denote this kind of learning as \emph{weak supervision}. We evaluate our discovered action classifiers on PASCAL VOC 2012 action classification dataset. As the training and test data are from two different domains (\emph{i.e.} \emph{domain transfer}), the differences may lead to dataset bias \cite{torralba2011unbiased}.

\textbf{Discussion:} The naive baseline is worse than our methods for two reasons: a) Humans might term the same action with different words, for example ``do flip" and ``jump". Searching keywords ignores all the other terms and their related images; b) An action might have multiple sub-categories: For example, ``play guitar" and ``play" piano are sub-categories of ``play instrument".  The problems of the naive approach highlights the value for our ACD method. Oquab \textit{et al} \cite{oquab2014learning} outperform our method (ACD-A) in mAP for 2 reasons: a) They use fully-labelled data and don't encounter domain differences: they finetune CNN on \emph{trainval} set of PASCAL VOC 2012 action classification. But we don't use any data from PASCAL VOC. b) They use person's bounding box in training and testing, but we don't. However, ACD-A still outperforms Oquab \textit{et al.}\cite{oquab2014learning} in ``play instrument" and ``use computer".

\section{Conclusions}
We presented a system to automatically discover, cluster and learn the action concepts from image-sentence corpora. Verb-Object pairs were extracted from sentences as action concept candidates. We used VGG features and 2 fold cross-validation to verify these candidates. Multimodal representation, which combines the visual and linguistic information, was adopted to represent and cluster the action concepts. More than 100 human action concepts and 81 action clusters were learned from source domain (Flickr30k). We applied the learned classifiers of action concept clusters to action recognition task. We also quantitatively evaluated the learned classifiers on a different transfer domain (PASCAL VOC 2012 action classification). To further improve the performance, AdaBoost was applied to combine multiple related concept classifiers for a single action tag. Finally, we showed promising results compared to previous fully supervised method. 



\section{Acknowledgments}
This research was supported, in part,
by the Office of Naval Research under grant N00014-13-1-0493.

%
\bibliographystyle{abbrv}
\bibliography{actionconcept}  
%
%
\end{document}